\documentclass[10pt,twocolumn,letterpaper]{article}

\usepackage{cvpr}              

\usepackage[dvipsnames]{xcolor}

\definecolor{cvprblue}{rgb}{0.21,0.49,0.74}
\usepackage[pagebackref,breaklinks,colorlinks,citecolor=cvprblue]{hyperref}
\usepackage{adjustbox}
\usepackage{multirow}
\usepackage{makecell}
\usepackage[dvipsnames]{xcolor}
\usepackage[accsupp]{axessibility} 
\def\paperID{15720} 
\def\confName{CVPR}
\def\confYear{2024}

\begin{document}
\title{BiTT: Bi-directional Texture Reconstruction of Interacting Two Hands from a Single Image}

\author{Minje Kim$^{1}$\\
$^1$ KAIST\\
\and
Tae-Kyun Kim$^{1,2}$\\
$^2$ Imperial College London\\
}

\twocolumn[{%
\renewcommand\twocolumn[1][]{#1}%
\maketitle
\begin{center}
\centering
\captionsetup{type=figure}
\includegraphics[width=\textwidth]{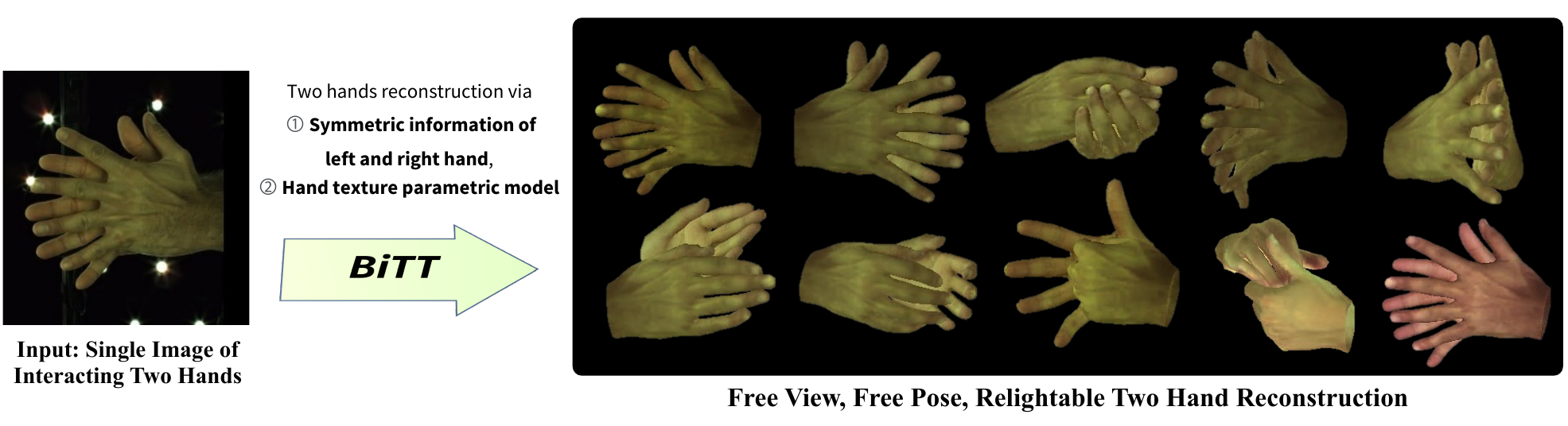}
\caption{Taking a single image input, our method renders the personalized texture of two hands at novel views, poses, and light conditions.}
\label{Teaser}
\end{center}
}]

\begin{abstract}
   Creating personalized hand avatars is important to offer a realistic experience to users on AR / VR platforms. While most prior studies focused on reconstructing 3D hand shapes, some recent work has tackled the reconstruction of hand textures on top of shapes. However, these methods are often limited to capturing pixels on the visible side of a hand, requiring diverse views of the hand in a video or multiple images as input. In this paper, we propose a novel method, BiTT(Bi-directional Texture reconstruction of Two hands), which is the first end-to-end trainable method for relightable, pose-free texture reconstruction of two interacting hands taking only a single RGB image, by three novel components: 1)\ bi-directional (left $\leftrightarrow$ right) texture reconstruction using the texture symmetry of left / right hands, 2) utilizing a texture parametric model for hand texture recovery, and 3)\ the overall coarse-to-fine stage pipeline for reconstructing personalized texture of two interacting hands. BiTT first estimates the scene light condition and albedo image from an input image, then reconstructs the texture of both hands through the texture parametric model and bi-directional texture reconstructor. In experiments using InterHand2.6M and RGB2Hands datasets, our method significantly outperforms state-of-the-art hand texture reconstruction methods quantitatively and qualitatively. The code is available at \url{https://github.com/yunminjin2/BiTT}.
\end{abstract}

\vspace{-1pt}
\section{Introduction}
\label{sec:intro}

3D human reconstruction has been studied in various areas. With the increasing usage of human-computer interaction, virtual reality (VR), and augmented reality (AR), reconstruction of human parts, including the full body, face, and hand, has been intensively studied for years. In particular, hand pose estimation, shape, and texture reconstruction are essential tasks for AR/VR interfaces. 3D hand reconstruction is still a challenging task due to the highly varied poses and shapes of hands. Previous works \cite{single1, single2, single3, single4, single5, single6} focused on estimating the 3D pose and shape of a single hand. A single hand reconstruction has been recently extended to two interacting hands \cite{IntagHand, Im2Hands, ACR, FourierHandFlow} and hand-object interaction \cite{HALO, GraspingField, WiYH, FirstPerson, MeasuringGeneralisation} scenarios.  

Learning the appearance of objects and humans is currently in active research for realistic reconstruction. NeRF \cite{NeRF} represents objects/scenes by a neural radiance field based on volume rendering. Appearance reconstruction of clothed human bodies \cite{Hvtr, Tava, ICON, NARF, PM3D} and faces \cite{face1, face2, face3, face4, derendering3} has been intensively studied compared to hands. For learning hand appearance, 
LISA \cite{LISA} used a radiance field to learn the shape and color appearance from multi-view images. In the latest works \cite{HandAvatar, HandNerf, LiveHand, HARP, LISA} to reconstruct the hand appearance, they take multi-view images or a monocular video as input to learn the texture of mostly single hands \cite{HandAvatar, HARP, LISA, LiveHand, HandNerf}. HandAvatar \cite{HandAvatar} and HARP \cite{HARP} render relightable hand appearance by estimating the albedo of a single hand.


3D reconstruction from a single image is also another challenging task. Non-visible side of an object should be estimated with given a single image for full 3D reconstruction. Self-Supervised 3D Mesh Reconstruction (SMR) \cite{SMR} estimates 3D meshes with texture from a single image in a self-supervised manner. Wu \textit{ et al.} \cite{vase} reconstructed vase artifacts into 3D mesh with environment lighting, shiny material, and texture albedo. Other works \cite{single1, single2, single3, single4} also focus on constructing symmetric objects or single-type objects. Human hands, however, exhibit non-symmetric features such as the palm and back. Given the information on one side of the hand, the other side of the hand texture should be estimated to fully reconstruct the hand. S2Hand \cite{S2Hand} and AMVUR \cite{AMVUR} presented a method to reconstruct both the appearance and geometry of a single hand from a single image. Nevertheless, their appearance is in blurred textures, omitting detail texture appearance.

\begin{figure}[!ht]
  \centering
  \includegraphics[width=8.3cm]{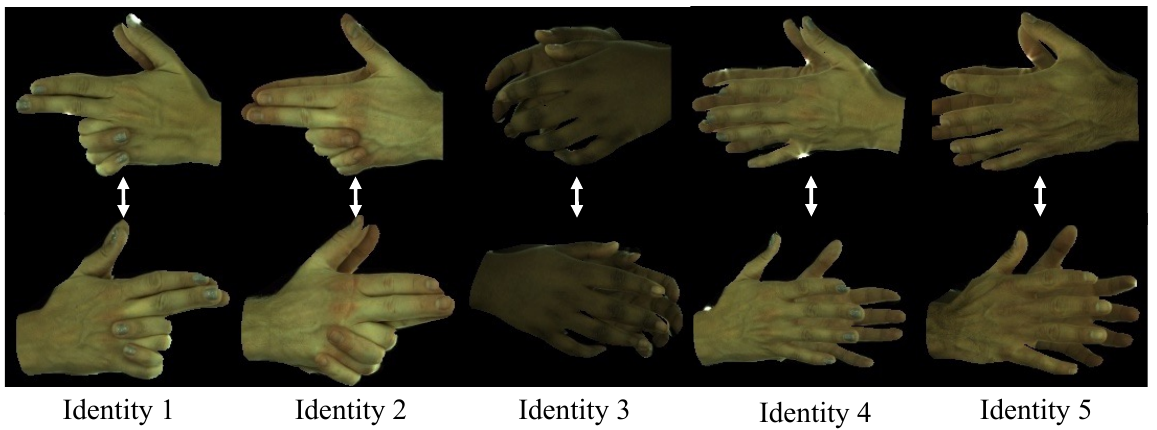}
  \caption{Symmetrical hand textures of different identities are shown, taken from pairs of diametrical camera views of InterHand2.6M \cite{InterHand2.6M}.}
  \label{Symmetry}
\end{figure}

In this paper, we propose a novel approach that exploits texture symmetry of left and right hands through the bi-directional reconstruction of two hand textures from a single image (see \cref{Symmetry}). Our method is trained per scene with only one single image, taking visible texture information from both hands and the parametric model of hand texture to reconstruct realistic hand appearances. Given an input image and a mesh of hands, our method predicts lights and albedo image (please refer to \cref{SE}). In the coarse stage, our model generates full hand textures with estimated vectors using HTML \cite{HTML}, the hand texture parametric model. With the estimated albedo image and coarse stage estimated texture map, the bi-directional texture reconstructor (BTR) yields the UV maps of both hand textures by utilizing the feature maps of a left and right hand. The proposed BiTT method can render both hands with fully controllable light conditions, poses, and camera views. We evaluated the method using the InterHand2.6M \cite{InterHand2.6M}, and RGB2Hands \cite{RGB2Hands} dataset and achieved high-fidelity appearances compared to state-of-the-art methods. We present qualitative results where the rendered images realistically capture personalized hand textures (e.g. wrinkles, veins, nails). \cref{Teaser} shows that our method is capable of controlling light conditions, camera views, and hand poses. To the best of our knowledge, this is the first method to reconstruct both hands with textures from a given single image input.


In summary, our main contributions are as follows: 1) we introduce a novel framework BiTT, the first method for rendering two interacting hands from a single image. 2) We propose the bi-directional texture reconstruction, exploiting the texture symmetry of left and right hands. 3) We introduce a way to use the texture parametric model for recovering invisible texture. 4) We demonstrate that our framework is an end-to-end trainable for photorealistic two-hand avatars with controllable poses, views, and light conditions. 
\vspace{-0.5cm}
\section{Related Work}
\label{RW}



\begin{figure*}[!ht]
  \centering
  \includegraphics[width=16.5cm]{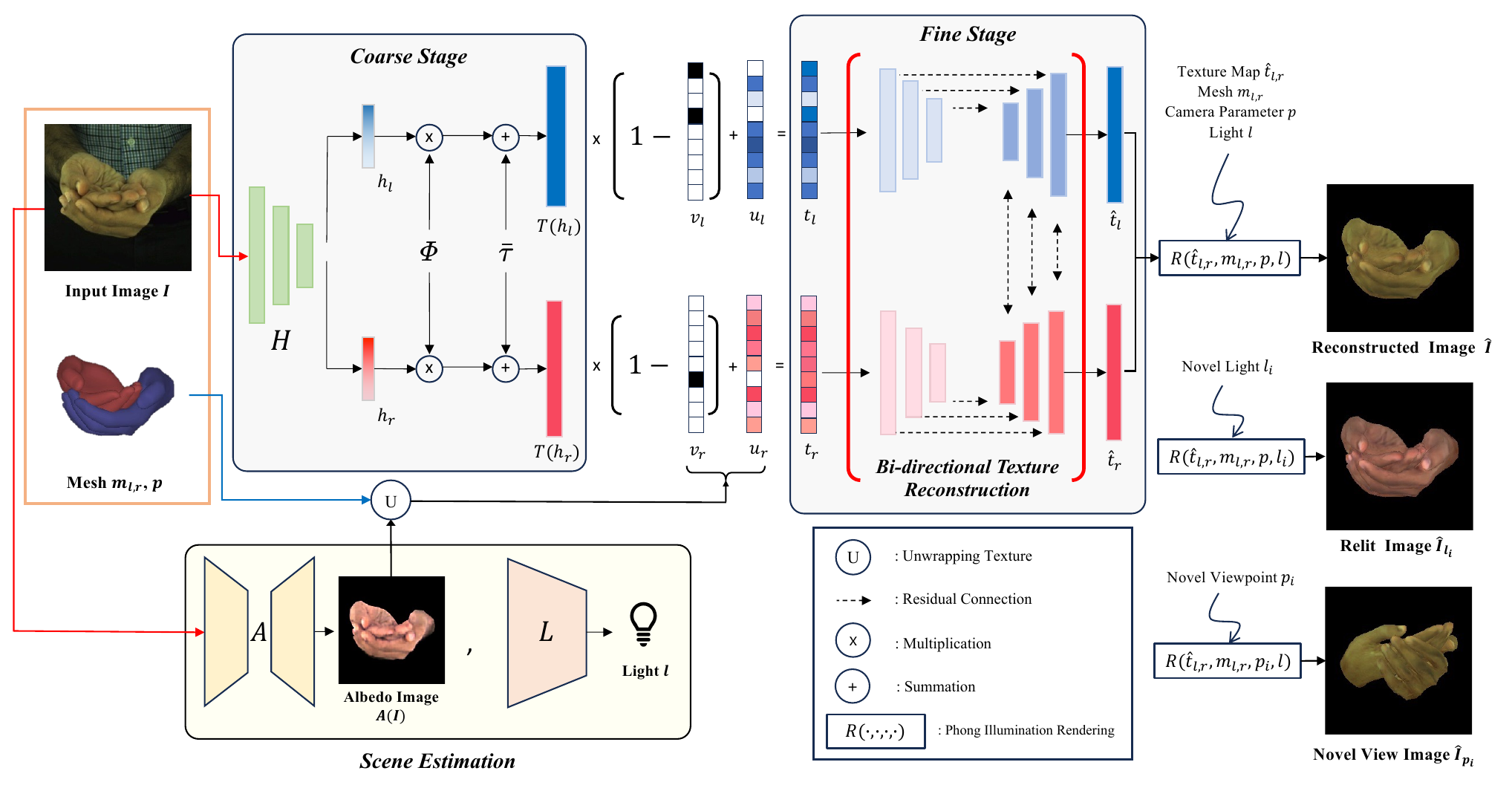}
  \caption{The architecture of BiTT. Our method consists of three steps: (1) scene estimation, (2) coarse stage, and (3) fine stage estimation. The scene estimation understands the scene by predicting the albedo image and lighting conditions with a given input image. Full detailed textures of both hands are reconstructed from the single image input. The hand texture parametric model is adopted in the coarse stage, then the bi-directional texture reconstruction refines the personalized hand textures by the texture symmetry of left-right-hands. 
  Finally, we render both hands with Phong Illumination \cite{phong}.}
  \label{Architecture}
\end{figure*}

In hand appearance modeling, NIMBLE \cite{NIMBLE} learns factorization of albedo, specular, and normal maps from high-definition hand textures. S2HAND \cite{S2Hand} estimates camera poses, colored meshes, and lighting conditions in a simultaneous way, but its rendering quality does not demonstrate detailed (or personalized) hand appearances. AMVUR \cite{AMVUR} reconstructs a hand using attention-based mesh vertices and the occlusion-aware texture regression model. However, its per-vertex texture reconstruction is not adaptable to the reconstruction of realistic hand textures due to its low resolution. In addition, the reconstructed hand textures contain background colors since the method does not consider geometric misalignment noise. LISA \cite{LISA} learns an implicit color field with implicit shape representation but still lacks detailed hand textures despite multi-view inputs.

\vspace{-0.3cm}
\paragraph{Neural-Radiance-Field based Representation.} In recent years, the application of hand representation by the neural radiance field has been studied extensively. HandAvatar \cite{HandAvatar} has improved the hand appearance with the occupancy field and predicted self-occlusion shadows. However, it requires a large amount of training data and time for a new instance of hand. HandNeRF \cite{HandNerf}, Livehand \cite{LiveHand} extracts hand texture from multi-view image sequence with volume rendering in neural radiance field for hand reconstruction. To develop subject-specific shape and appearance using NeRF-based methods such as HandAvatar, HandNeRF, and LiveHand, a substantial volume of images within a fixed light condition, up to thousands of images for each single sequence, is typically needed for training. Since these methods are integrated into an implicit space, they require additional steps for mesh extraction to further applications. They also face difficulties in controlling illumination, including rendering shadows and editing textures.

\paragraph{UV Texture Map based Representation.} Unlike the implicit function-based methods, HARP \cite{HARP} reconstructed a single hand by mesh rendering with a UV texture map. HARP reconstructs a personalized single hand using the UV texture map, normal map, and self-occlusion shadow. HARP renders detailed hand textures including out-of-distribution appearance like tattoos and accessories. However, the HARP method assumes the texture color into a specific value, and as a result, the occluded part of the texture tends to be a fixed value, missing out the detailed texture.

\paragraph{Positioning BiTT to w.r.t related works.} Two-hand texture rendering based on UV maps has not been explored to the best of our knowledge. In implicit space, HandNeRF \cite{HandNerf} reconstructed two hands from a multi-view image sequence. Reconstruction of interacting two hands from a single image is a challenging task, as we have to lift a 2D image to achieve accurate 3D hand representation. Also, two hands encounter a higher rate of occlusion compared to a single hand, thus restoring occluded texture is another challenge. Whereas using multi-views or a video sequence as input helps reconstruct two interacting hands, taking a single image input poses more significant challenges. However, two hands convey more information than a single hand, which enables us to reconstruct both hands realistically within a single image. In this work, we propose a novel method that exploits the two-hand symmetric texture information and employs the hand texture parametric model as prior. The proposed method demonstrates that only a single image is enough to reconstruct a realistic two-hand avatar in contrast to the prior works \cite{HARP, HandAvatar, LiveHand, HandNerf}.


\section{Methods}

We propose a texture reconstruction model for two interacting hands with a single image. \cref{Architecture} shows the architecture of BiTT. Given a single RGB image $I$ of interacting two hands, we reconstruct realistic personalized textures for a two-hand avatar. BiTT is composed of three steps: the scene estimation, the coarse stage based on the parametric model, and the fine stage composed with the bi-directional texture reconstructor.

For reconstructing the full texture of both hands from a single image, our method relies on utilizing the symmetry between the left and right hands. While the disparity between the left and right hand texture is assumed to be not significant (e.g. see \cref{Symmetry}), it is beneficial to leverage the symmetrical data to reconstruct detailed textures even on occluded hands. For those pixels not visible on both hands, the texture parametric model is adopted, and its estimation is further refined. 

\subsection{Scene Estimation}
\label{SE}
Scene estimation involves estimating the environment, including the light and albedo of hands in the input image. To estimate the environment, our light network $L$  estimates the light parameter $l$ which is composed of the ambient light color, diffuse, specular, and direction. Furthermore, we also predict an albedo image through our albedo network $A$, which represents the surface reflectance of objects using the Lambertian surface. Details of the light and albedo networks are found in the supplementary material. 

\subsection{Coarse Stage}
\label{CS}
Since 2D image lacks the information to reconstruct 3D objects, non-symmetric object reconstruction \cite{PIFU, ICON, PM3D, RealFusion, Anything3D, S2Hand, TextureField} suffers from blurry textures. Several major works \cite{HandAvatar, HARP, NeRF, LISA, LiveHand} overcome the lack of information by multi-view images or a monocular video as input. Instead of increasing the input information, we augment data through the parametric model. The coarse stage is based on the hand texture parametric model (HTML) \cite{HTML}, which encodes the hand texture through PCA algorithm and is able to create a full texture of hand from a single image. 

\paragraph{Background on HTML \cite{HTML}.}
HTML is a hand texture parametric model that can create a full-texture UV map with a given vector. HTML scanned 51 hands and aligned them to a canonical space with MANO \cite{MANO} model fitting. After the MANO fitting, texture mapping is done with manually defined UV coordinates. Finally, a parametric model $T$ is created by PCA on vectors $\tau_i \in \mathbb{R}^{618990}$ with the collection of 2D texture maps where 618,990 is a total number of 206,330 pixels in texture map with RGB channels. Given the covariance matrix $C \in \mathbb{R}^{618990 \times 618990} = \frac{1}{n-1}\sum^{n}_{i=1}{(\tau_i-\bar{\tau})(\tau_i-\bar{\tau})^\top}$, where $\bar{\tau} = \frac{1}{n}\sum^{n}_{i=1}{\tau_i}$, the principal components $\mathit{\Phi} \in \mathbb{R}^{618990 \times 101}$ are obtained by singular value decomposition of $C = \mathit{\Phi} \Sigma \mathit{\Phi}^{\top}$, where $\Sigma \in \mathbb{R}^{101 \times 101}$ is a diagonal matrix. With the principal components $\mathit{\Phi}$, we get a full texture eigenvalue $T(\alpha) \in \mathbb{R}^{618990}$ for a given parameter vector $\alpha \in \mathbb{R} ^ {101}$ with $ T(\alpha) = \bar{\tau} + \mathit{\Phi} \alpha$. For more details on HTML, please refer to \cite{HTML}.

\paragraph{Hand Texture Parametric Model in Coarse Stage.}
The HTML \cite{HTML} network $H$ in \cref{Architecture} is an encoder that estimates both the hand parameter vectors from an input image $I$. Given the left-hand parameter vector $h_l$ and the right-hand parameter vector $h_r$, we can obtain the full texture eigenvalues $T(h_l), T(h_r)$. It can be formally defined as:

\begin{equation} \label{LossFunction1}
    T(h_i) = \bar{\tau} + \mathit{\Phi} h_i \textrm{, where } h_l,h_r = H(I), i=l,r
\end{equation} 
\begin{equation}
\label{LossFunction1a}
    \hat{I}_{coarse} = \left\{  R(T(h_i), m_i, p, l)\right\}_{i=l,r} 
\end{equation}

$I$ is an input image, $R$ is the differentiable renderer based on Phong model \cite{phong}, and $\hat{I}_{coarse}$ is the reconstructed image from the coarse stage. With the texture vector $T(h_i)$ and meshes $m_i$, $R$ renders two hand meshes with texture on the 2D space at a camera viewpoint $p$ and light condition $l$.

\subsection{Fine Stage}
\begin{figure}[!ht]
  \centering
  \includegraphics[width=8.1cm]{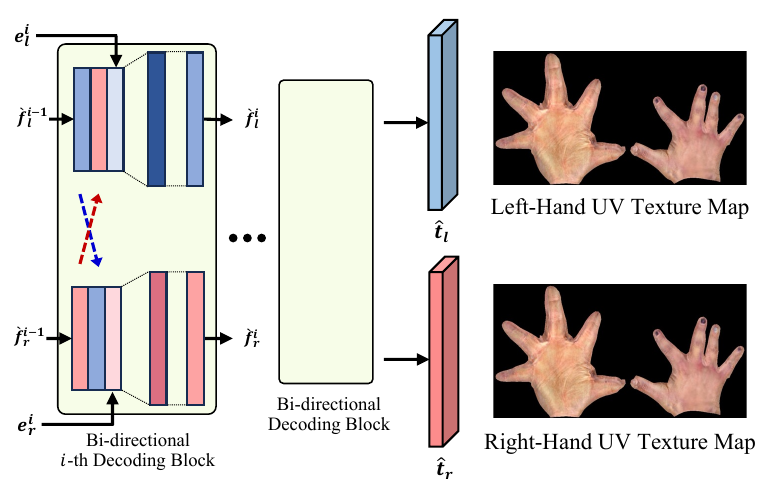}
  \caption{Detailed architecture of the decoding layer in the bi-directional texture reconstruction. }
  \label{Detail}
\end{figure}
\label{FS}
 To render more realistic and personalized textures, we use the symmetric features of left/right hands rather than using features independently. We propose a novel bi-directional texture reconstructor (BTR) to efficiently use the symmetric texture features of both hands. BTR reconstructs the full texture from visible pixels in the albedo image $A(I)$. We create a visible UV texture map and texture vector by unwrapping hand textures as we know the mapping between the UV texture map and image pixels, as well as the mapping between the UV texture map and the texture vector. With the visible texture vector for each hand unwrapped from the albedo image $A(I)$, denoted as $u_l, u_r$ for each hand, and $v_l, v_r$ which are visible mask texture vectors for each hand, we synthesize texture vector $t_l, t_r$ defined as:
 \begin{equation} \label{LossFunction2}
    t_i = T(h_i) (1 - v_i) + u_i \textrm{, where } i=l,r
\end{equation} 
BiTT, then, generates full both hand textures through the BTR.

In addition, with the visible albedo texture $u_l, u_r$, we render $\hat{I}_{albedo}$ with the estimated light $l$  to use in loss function:
\begin{equation}
\label{LossFunction1b}
    \hat{I}_{albedo} = \left\{  R(u_i, m_i, p, l)\right\}_{i=l,r} 
\end{equation}

\paragraph{Bi-directional Texture Reconstruction.} 
Given the synthesized texture vectors $t_l, t_r$, the bi-directional texture reconstructor (BTR) encodes each texture vector, denoted as $e_l, e_r$. After encoding each hand texture vector, the decoder in BTR decodes the embedded feature to create a full texture vector using skip connections. \cref{Detail} describes the detailed description of the BTR decoder. In each bi-directional decoding block, the embedded features of the same hand are concatenated with a skip connection. We also concatenate the other hand texture embedded features in a bi-directional way to use the symmetric information of two hands. Afterward, we obtain the decoded variable from the concatenated feature at the i-th level, denoted as $\grave{f}^{i}$. The decoded features $\grave{f}^{i}_{l}, \grave{f}^{i}_{r}$, which means the $i$-th level of the left hand and the right hand, respectively, are defined as:
\begin{equation} \label{LossFunction3}
    \grave{f}^{i}_{l} = \sigma(\mathcal{N}([e^{i}_{l}, \grave{f}^{i-1}_{l}, \grave{f}^{i-1}_{r}]))
\end{equation} 
\begin{equation} \label{LossFunction4}
    \grave{f}^{i}_{r} = \sigma(\mathcal{N}([e^{i}_{r}, \grave{f}^{i-1}_{r}, \grave{f}^{i-1}_l]))
\end{equation} 
where $\sigma$ denotes a ReLU activation function, $\mathcal{N}$ is a convolution neural network (CNN), and [] denotes the channel concatenation operator.

After the decoding, we obtain the full texture vectors $\hat{t_l}, \hat{t_r} \in \mathbb{R}^{3 \times 206330}$ for each hand. Thus, our network is formulated as:
\begin{equation} \label{parametric_texture_eq1}
    \hat{I} = BiTT(I) = \left\{R(\hat{t}_i, m_i, p, l) \right\}_{i=l,r}
\end{equation} 
where $\hat{I}$ is the final rendered image of reconstructed both hands.


\begin{table*}[t]
\centering
\caption{Quantitative comparisons of BiTT, S2Hand \cite{S2Hand}, HTML \cite{HTML} and HARP \cite{HARP}.  Training data are from Interhand2.6M \cite{InterHand2.6M} including all identities. For evaluations, novel poses and viewpoints are randomly selected from the same hand identity.  In the case when not using GT mesh, we used IntagHand \cite{IntagHand} for obtaining meshes. }
\label{table1}
\begin{subtable}{0.49\textwidth}
\raggedright{
\caption{Using GT mesh in all methods.}
\label{table1:a}
\begin{adjustbox}{scale=0.76}
\begin{tabular}{@{}cccccc@{}}
\toprule
Evaluation                                                      & Method              & L1↓             & LPIPS↓          & PSNR↑          & MS-SSIM↑           \\ \midrule
\multicolumn{1}{c|}{\multirow{3}{*}{\makecell{Appearance \\ Reconstruction}}} & S2Hand \cite{S2Hand}             & 0.0206          &  0.1340          & 26.39           & 0.8570              \\
\multicolumn{1}{c|}{}                                           & HTML \cite{HTML}    & 0.0256          & 0.1292          & 24.72          & 0.8152     \\ 
\multicolumn{1}{c|}{}                                           & HARP \cite{HARP} & 0.0157      & \textbf{0.0696}       & 28.11       & 0.9061             \\     \cmidrule(l){2-6} 
\multicolumn{1}{c|}{}                                           & \textbf{BiTT(ours)} &      \textbf{0.0101}       & 0.1019       & \textbf{30.41}       & \textbf{0.9349}    \\ \midrule
\multicolumn{1}{c|}{\multirow{3}{*}{Novel Poses}}               & S2Hand            &  0.0221          & 0.1343          & 25.70          & 0.8507           \\
\multicolumn{1}{c|}{}                                           & HTML    & 0.0255         & 0.1291          & 24.49          & 0.8153          \\ 
\multicolumn{1}{c|}{}                                           & HARP & 0.0239      & 0.1266       &    25.79       &        0.8546     \\     \cmidrule(l){2-6} 
\multicolumn{1}{c|}{}                                           & \textbf{BiTT(ours)} & \textbf{0.0209}      & \textbf{0.1261}      & \textbf{26.54}       & \textbf{0.8564}      \\ \midrule
\multicolumn{1}{c|}{\multirow{3}{*}{Different Views}}            & S2Hand            & 0.0217          & 0.1320          & 25.73          & 0.8484            \\
\multicolumn{1}{c|}{}                                           & HTML    & 0.0254          & 0.1282          & 24.42          & 0.8133             \\  
\multicolumn{1}{c|}{}                                           & HARP     & 0.0234      &  0.1189       & 25.97       &  0.8346   \\  \cmidrule(l){2-6} 
\multicolumn{1}{c|}{}                                           & \textbf{BiTT(ours)} & \textbf{0.0204}      & \textbf{0.1092}              & \textbf{27.79}       & \textbf{0.8843}      \\ \bottomrule
\end{tabular}
\end{adjustbox}}
\end{subtable}
\begin{subtable}{0.49\textwidth}
\raggedleft{
\caption{Without using GT mesh in all methods. }
\label{table1b}
\begin{adjustbox}{scale=0.76}
\begin{tabular}{@{}cccccc@{}}
\toprule
Evaluation                                                      & Method              & L1↓             & LPIPS↓          & PSNR↑          & MS-SSIM↑           \\ \midrule
\multicolumn{1}{c|}{\multirow{3}{*}{\makecell{Appearance \\ Reconstruction}}} & S2Hand \cite{S2Hand}       & 0.0264          &  0.1214          & 25.72           & 0.8897      \\
\multicolumn{1}{c|}{}                                           & HTML \cite{HTML}    & 0.0268          & 0.1207          & 24.48          & 0.8545     \\ 
\multicolumn{1}{c|}{}                                           & HARP \cite{HARP} & 0.0237      & 0.1047       & 25.17       & 0.8697          \\     \cmidrule(l){2-6} 
\multicolumn{1}{c|}{}                                           & \textbf{BiTT(ours)} &      \textbf{0.0131}       & \textbf{0.1044}       & \textbf{28.40}       & \textbf{0.9093}    \\ \midrule
\multicolumn{1}{c|}{\multirow{3}{*}{Novel Poses}}               & S2Hand            &  0.0280          & 0.1525          & 23.06          & 0.8092           \\
\multicolumn{1}{c|}{}                                           & HTML    & 0.0310         & 0.1299          & 23.46          & 0.8281          \\ 
\multicolumn{1}{c|}{}                                           & HARP & 0.0256        & 0.1410       &    24.32       &     0.8419     \\     \cmidrule(l){2-6} 
\multicolumn{1}{c|}{}                                           & \textbf{BiTT(ours)} & \textbf{0.0223}      & \textbf{0.1228}      & \textbf{25.12}       & \textbf{0.8423}      \\ \midrule
\multicolumn{1}{c|}{\multirow{3}{*}{Different Views}}           & S2Hand            & 0.0244          & 0.1512          & 24.22          & 0.8335   \\
\multicolumn{1}{c|}{}                                           & HTML    & 0.0291          & 0.1297          & 24.22          & 0.8375             \\  
\multicolumn{1}{c|}{}                                           & HARP     & 0.0251      &  0.1367       & 24.49       &  0.8507   \\  \cmidrule(l){2-6} 
\multicolumn{1}{c|}{}                                           & \textbf{BiTT(ours)} & \textbf{0.0210}      & \textbf{0.1273}              & \textbf{26.34}       & \textbf{0.8674}      \\ \bottomrule
\end{tabular}
\end{adjustbox}}
\end{subtable}
\end{table*}

\begin{table*}[t]
\centering
\caption{Quantitative comparisons between compared methods in RGB2Hands \cite{RGB2Hands} dataset. All training and testing images are randomly selected. As RGB2Hands has no ground truth mesh, we used IntagHand \cite{IntagHand} as an off-the-shelf model for two hand mesh reconstruction.}
\label{table2}
\begin{adjustbox}{scale=0.82}
\begin{tabular}{@{}ccccccc@{}}
\toprule
Evaluation                                                      & Method              & L1↓             & LPIPS↓          & PSNR↑          & SSIM↑           & MS-SSIM↑        \\ \midrule
\multicolumn{1}{c|}{\multirow{4}{*}{Appearance Reconstruction}} & S2Hand \cite{S2Hand}     & 0.0179          & 0.0601          & 25.72           & 0.9459         & 0.9286          \\
\multicolumn{1}{c|}{}                                           & HTML \cite{HTML}    & 0.0203          & 0.0923          & 24.42          & 0.8927          & 0.9075          \\ 
\multicolumn{1}{c|}{}                                           & HARP \cite{HARP} &  0.0155      & \textbf{0.0433}       & 25.63       & 0.9309      & \textbf{0.9344} \\     \cmidrule(l){2-7} 
\multicolumn{1}{c|}{}                                           & \textbf{BiTT(ours)} & \textbf{0.0148}      & 0.0683       & \textbf{26.02}       & \textbf{0.9501}      & 0.9323 \\ \midrule
\multicolumn{1}{c|}{\multirow{4}{*}{Novel Poses}}               & S2Hand            & 0.0222          & 0.0778          & 24.22          & 0.9326          & 0.8991         \\
\multicolumn{1}{c|}{}                                           & HTML    & 0.0233          & 0.0961          & 23.25         & 0.8829          & 0.8900         \\ 
\multicolumn{1}{c|}{}                                           & HARP & 0.0208      & \textbf{0.0758}       & 23.88       & 0.9043      & 0.9042 \\    \cmidrule(l){2-7} 
\multicolumn{1}{c|}{}                                           & \textbf{BiTT(ours)} & \textbf{0.0196}      & 0.0774      & \textbf{24.54}       & \textbf{0.9352}      & \textbf{0.9046} \\ \bottomrule
\end{tabular}
\end{adjustbox}
\end{table*}

\subsection{Loss Functions}
The loss functions we use to train our model are as follows.

\paragraph{Reconstruction Loss.}
As our method aims to represent the realistic appearance of the input image, we included the reconstruction loss $\mathcal{L}_{rec}$ to measure the similarity between the input image $I$ and three distinct rendered images: rendered image from fine stage ($\hat{I}$), rendered image from coarse stage ($\hat{I}_{coarse}$), and the rendered image with albedo visible texture ($\hat{I}_{albedo}$). The reconstruction loss  $\mathcal{L}_{rec}$ is defined as: 
\begin{multline}
\label{LossFunction5}
    \mathcal{L}_{rec} =  \lambda_{rec}\| (I - \hat{I})\|_1 + \lambda^{coarse}_{rec}\| I - \hat{I}_{coarse}\|_1 \\ 
    + \lambda^{albedo}_{rec}\| I - \hat{I}_{albedo}\|_1 
\end{multline}

\paragraph{Reconstruction Loss on Non-visible Pixels.}
Visible information itself is not enough to accurately recreate the complete texture of hands. Even when symmetrical aspects are taken into account, it is still not sufficient to cover the entire hand texture. Thereby, we resort to the full texture obtained in the coarse stage for those pixels not observed in either of the hands. We measure the L1 loss between the coarse stage estimated hand texture and fine stage estimated hand texture with the invisible map mask. $\mathcal{L}_{nv}$ is defined as:
\begin{equation} 
\label{nv_loss}
    \mathcal{L}_{nv} = \sum_{i=l,r}{\|((T(h_i) - \hat{t}_i) (1-v_i) \|_1}
\end{equation} 
where $T(h_i)$ is the reconstructed hand texture in the coarse stage, $\hat{t}_i$ is the reconstructed hand texture in the fine stage for each hand. The visible mask of the hand texture is denoted as $v_i$ and thus, $1 - v_i$ represents the invisible mask of the texture. These textures are simultaneously refined with those of visible and symmetric texture reconstruction and consistency losses. 

\paragraph{Albedo Consistency Loss.} The albedo image should be obtained independent of lighting conditions. To obtain an accurate albedo image, we augment individually rendered images $\hat{I}_{coarse, \hat{l}_i}$ with different lighting conditions $\hat{l}_i$ and obtain the albedo images through the albedo network $A$. Our new albedo loss function $\mathcal{L}_{alb}$ calculates the L1 loss between albedo images from different lights. Thus, the albedo loss term $\mathcal{L}_{alb}$ is defined as:
\begin{equation} 
\label{albedo}
    \mathcal{L}_{alb} = \sum^n_{i=1}{\sum^n_{j=i+1}{ [ \|A(\hat{I}_{coarse, \hat{l}_i}) - A(\hat{I}_{coarse, \hat{l}_j}) \|_1]}}. 
\end{equation} 
In our experiment, we rendered three different lights in total: a reconstructed light, a light from a different direction, and a light with a different color to make the albedo network $A$ estimate the albedo image consistently even in the different lighting conditions. 

\paragraph{Symmetric Loss.}
 For learning the symmetric feature of two hands, we apply a symmetric loss term $\mathcal{L}_{sym}$ which is the L1 loss between the left and right hands. The $\mathcal{L}_{sym}$ is defined as:
\begin{equation} \label{SymLoss}
    \mathcal{L}_{sym} = \lambda_{sym}\| (\hat{t}_l - \hat{t}_r)\|_1
\end{equation} 

\paragraph{Total Loss.}
In summary, our loss function is defined as:
\begin{equation} \label{LossFunction}
    \mathcal{L} = \mathcal{L}_{rec} + \lambda_{nv}\mathcal{L}_{nv} + \lambda_{alb}\mathcal{L}_{alb} + \lambda_{sym}\mathcal{L}_{sym}
\end{equation} 
where $\lambda^{coarse}_{rec} = 0.8$, $\lambda^{albedo}_{rec} = 0.4$, $\lambda_{nv} = 0.2$, $\lambda_{alb} = 0.2$, $\lambda_{sym} = 0.3$  are used to train our model and fixed for all experiments. 

\section{Experiments}
\subsection{Experimental Settings}
Our training framework follows a per-scene training like NeRF \cite{NeRF}. Given multiple images of the scene, they learn to generate novel views of the scene. Efforts have been made to reduce the number of required images, such as PixelNeRF \cite{PixelNeRF} which trains NeRF with just one or a few images. In addition, HARP \cite{HARP}, HandAvatar \cite{HandAvatar}, HandNeRF \cite{HandNerf}, and others \cite{LiveHand, LISA} are based on per-scene training, requiring dozens of multiple frames of the same hand (scene). Our method, utilizing texture symmetry and a parametric texture model, requires only a single image for training.

Our model is trained in end-to-end manner. For each scene, our training process involves 700 epochs with a learning rate decay by half every 200 steps, starting from an initial rate of 0.001. Each scene training process is completed in less than 7 minutes.

\paragraph{Datasets.}
We mainly use the InterHand2.6M \cite{InterHand2.6M} dataset which is composed of interacting two hands for both quantitative and qualitative evaluations. Since HARP \cite{HARP} and NeRF-based approaches \cite{HandAvatar, HandNerf, LiveHand} require hundreds of images per scene for training, can only involve a few scenes to experiments. Requiring a single image, our experiments are conducted through a total of 378 scenes (images) utilizing all 26 hand identities of the dataset. After training, we evaluated its performance with different poses and different views of the hands from the same identity. Specifically, we evaluated with 8 distinct poses and 8 varying viewpoints for each scene, resulting in a total of 6,048 testing images. 

We also evaluated our model with the RGB2Hands \cite{RGB2Hands} dataset, however, since RGB2Hands does not present multiview images, we evaluated it with different pose images. Across 300 scenes in the RGB2Hands dataset, we trained our method and evaluated it with 40 different poses of images for each scene, thus 12,000 testing images in total.

For each dataset, we present the proportion of visible, invisible, and usable symmetric texture pixels in each hand texture at \cref{table3}. Using symmetric information, we can acquire up to 60\% texture of each hand; otherwise, we have only about 35\% full texture available from the input images. This demonstrates that using the symmetric information between both hands is reasonable for reconstructing two hand textures from a single image.


\paragraph{Metrics.} 
For the quantitative comparisons of different appearance models, we use a set of metrics that is often applied to assess the fidelity and quality of rendered images. We use the L1, learned perceptual image patch similarity (LPIPS) \cite{LPIPS}, the structural similarity metric (SSIM) \cite{SSIM}, the multiscale structural similarity metric (MS-SSIM) \cite{MS-SSIM}, and the peak signal-to-noise ratio (PSNR).

\begin{table}[t]
\centering
\caption{Pixel ratio comparison between left-hand, right-hand visible texture, usable symmetric texture, and invisible texture in our experiment dataset. Colors refer to the \cref{UV_seg} mask label.}
\label{table3}
\begin{adjustbox}{scale=0.8}
\begin{tabular}{@{}c|c|c@{}}
\toprule
Dataset                    & InterHand2.6M \cite{InterHand2.6M} & RGB2Hands \cite{RGB2Hands} \\ \midrule
\textcolor{RoyalBlue}{Left-hand visible texture}  & 35.40\%       & 39.72\%   \\
\textcolor{red}{Right-hand visible texture} & 36.44\%       & 39.06\%   \\
\textcolor{Goldenrod}{Usable symmetric texture}   & 24.29\%       & 10.21\%   \\
\textcolor{gray}{Invisible texture}          & 19.89\%       & 24.95\%   \\ \bottomrule
\end{tabular}
\end{adjustbox}
\end{table}

\begin{figure}[t]
  \centering
  \includegraphics[width=8.0cm]{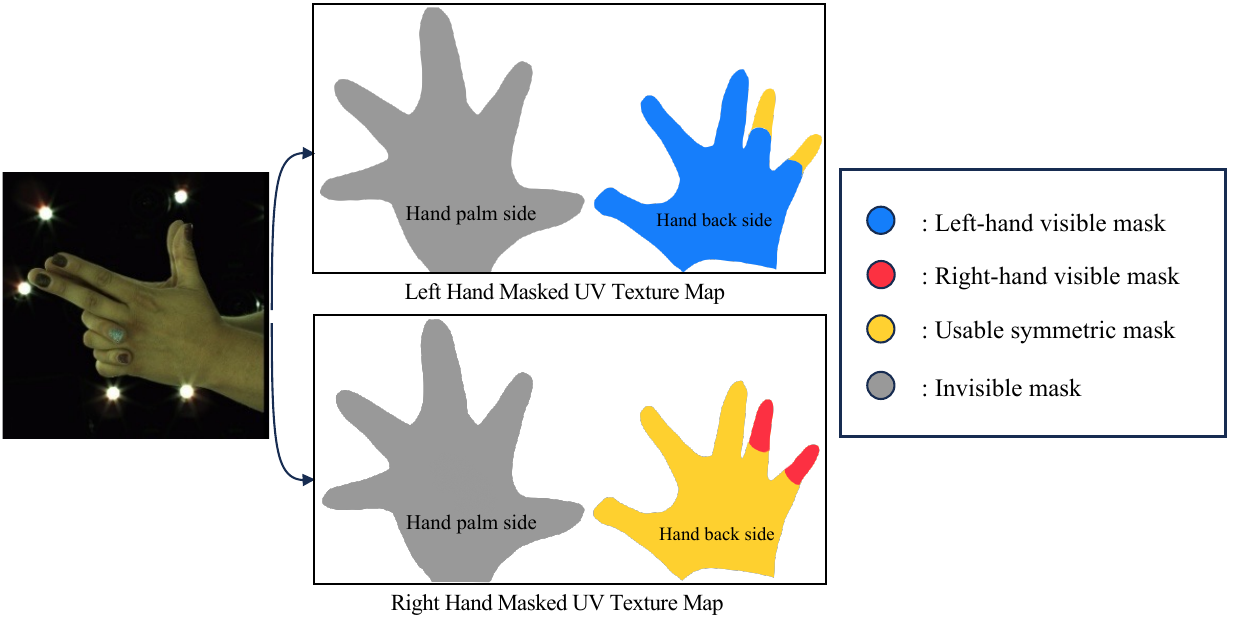}
  \caption{This figure shows the visible, invisible, usable symmetric texture mask on the UV texture map from an image. }
  \label{UV_seg}
\end{figure}

\paragraph{Compared Methods.}
We compare our method with S2Hand \cite{S2Hand} which reconstructs a single hand with per-vertex texture rendering. HARP \cite{HARP} and HTML \cite{HTML} are the methods that reconstruct a single hand with UV map rendering, and we compare the appearance quality among these methods. We modify S2Hand, HTML, and HARP to two hands for comparison. These methods are extended to estimate each hand texture discretely while learning each hand texture from a rendered image. Implicit function-based methods \cite{HandAvatar, HandNerf, LISA, LiveHand} are not included in the comparison, as their methods are not straightforward to extend for two hands, and their methods require at least hundreds of images per scene for training. For showing robustness on geometric misalignment, we present two result tables; using ground truth mesh in all the compared methods at \cref{table1:a} and without using ground truth mesh in all methods at \cref{table1b}. Where the ground truth mesh is unavailable, such as RGB2Hands data, we initialize hand meshes using the off-the-shelf method, IntagHand \cite{IntagHand}. 

\begin{figure}[t]
  \centering
  \includegraphics[width=8.2cm]{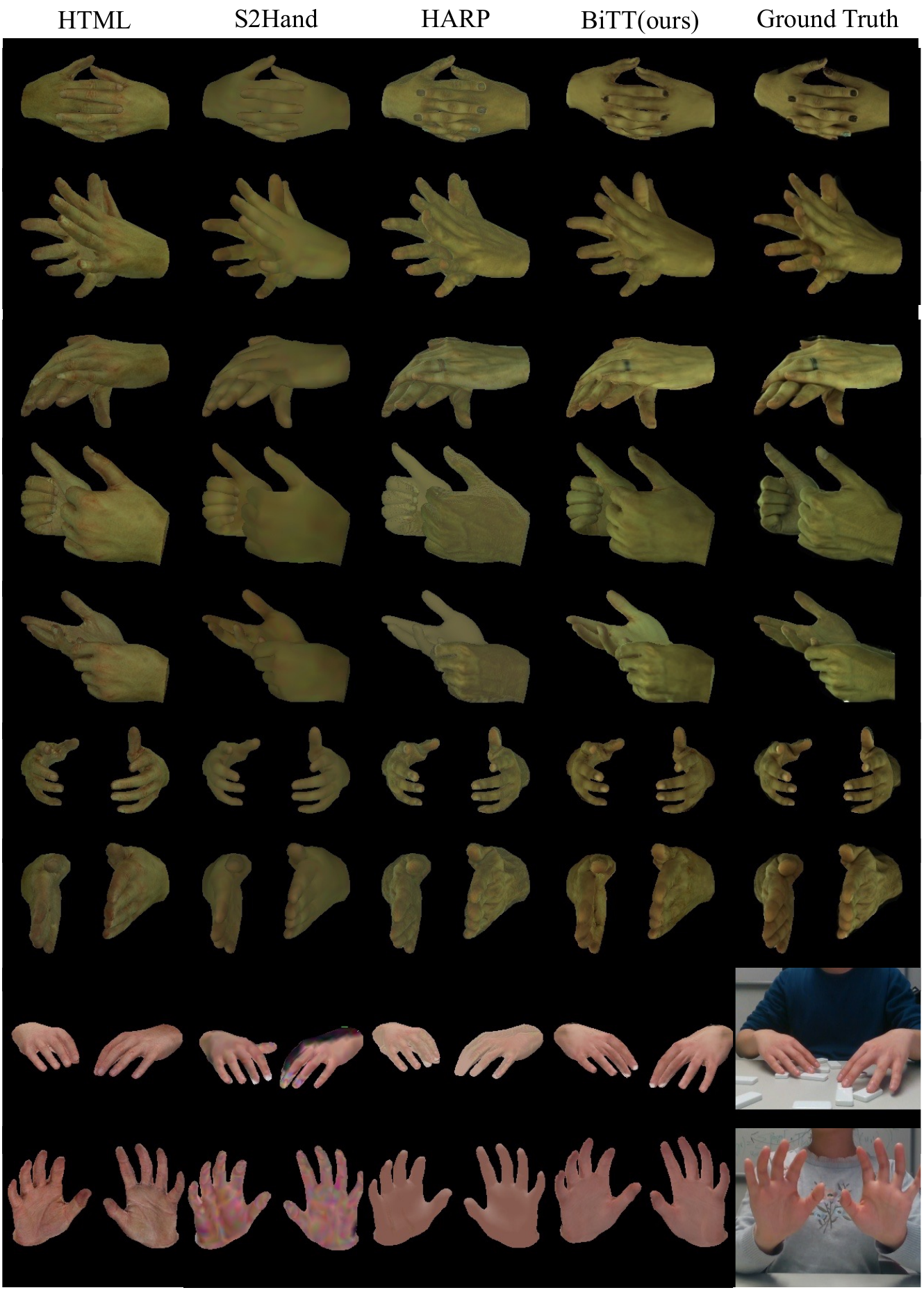}
  \caption{Qualitative results of HTML \cite{HTML}, S2Hand \cite{S2Hand}, HARP \cite{HARP}, and BiTT rendered on novel-pose and viewpoint. The last two rows pertain to the RGB2Hands \cite{RGB2Hands} dataset, while the remaining rows are from the InterHand2.6M \cite{InterHand2.6M} dataset.}
  \label{Qualitative}
\end{figure}

\begin{figure*}[t]
\centering
\includegraphics[width=17.1cm]{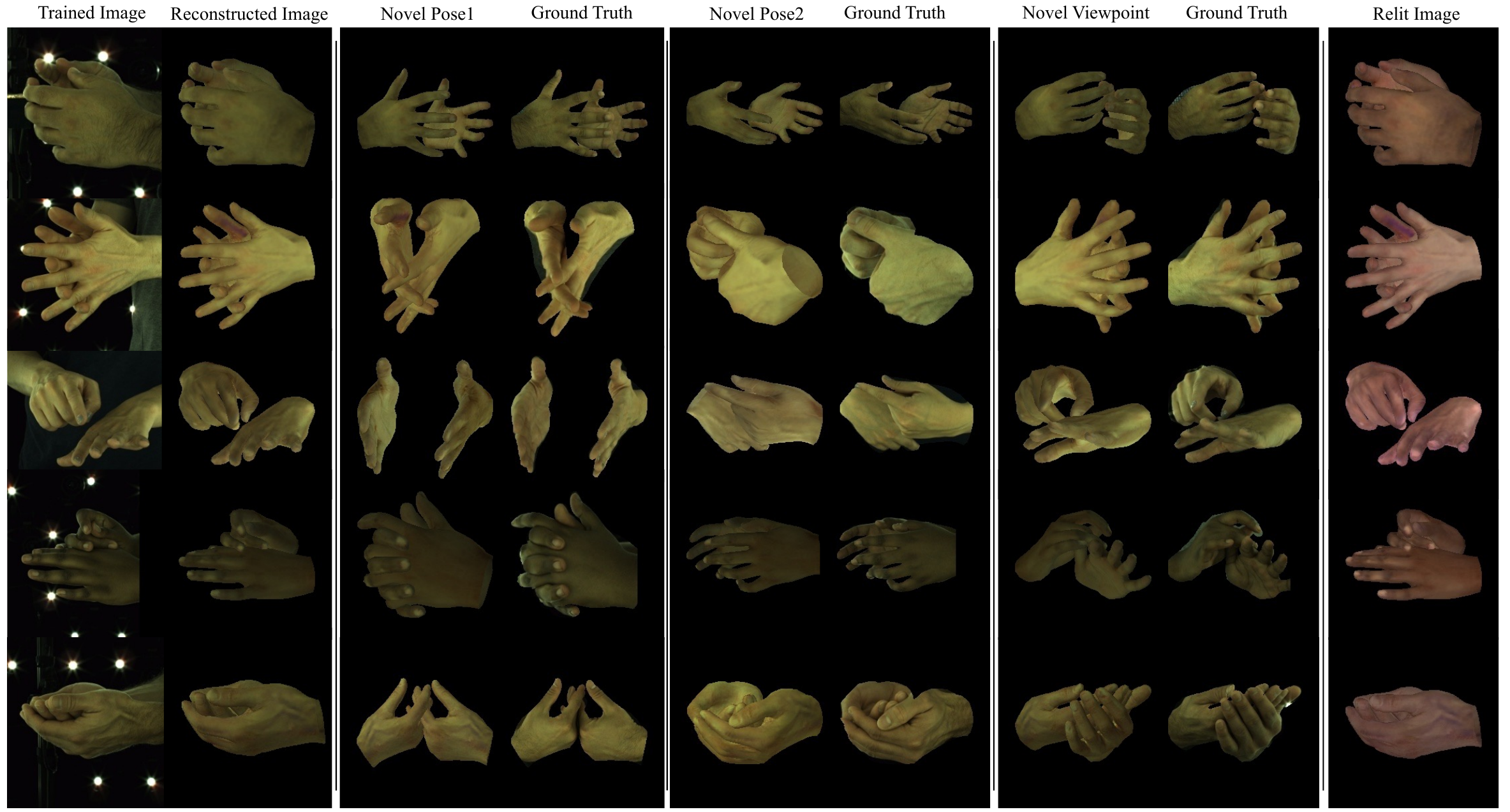}

\caption{Using only a single image input, our method reconstructs realistic detailed textures for both hands. We present some results having different poses, viewpoints with corresponding ground truth images, and relighted images.}

\label{MoreResult}
\end{figure*}

\subsection{Evaluation on Texture Reconstruction}
\paragraph{Comparing with Prior Arts.}
We show the qualitative results in \cref{Qualitative} and the quantitative comparisons in \cref{table1} and \cref{table2}. The results show that BiTT significantly outperforms other baselines, especially in reconstructing the invisible hand parts. HTML \cite{HTML}, S2Hand \cite{S2Hand} are not able to represent detailed appearances like vessels, wrinkles, and hair, whereas BiTT captures detailed personalized appearances using symmetric information. HARP \cite{HARP} excels in the visible side appearance (regarding \cref{table1}, \cref{table2}), but it lacks the capability to reconstruct the invisible sides, merely displaying a uniform color in those areas. Notably, BiTT remains robust to geometric misalignments, maintaining high performance even without using GT as having the advantages of the parametric model. The performance gain on RGB2Hands \cref{table2} is relatively less significant than on InterHand2.6M \cite{InterHand2.6M}. This is due to the fact that RGB2Hands exhibits a lower percentage of usable symmetric texture, indicated in \cref{table3}. More results of BiTT are shown at \cref{MoreResult}. 

\begin{figure}[t]
\centering
  \includegraphics[width=8.2cm]{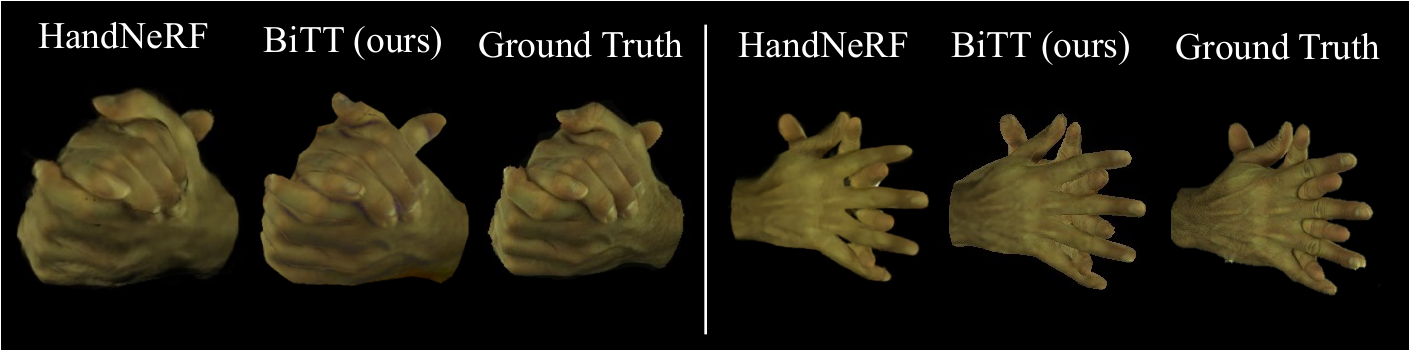}
  \caption{Qualitative results of HandNeRF \cite{HandNerf} and BiTT.}
  \vspace{-0.1cm}
  \label{HandNeRF_compare}
\end{figure}

We qualitatively compare HandNeRF \cite{HandNerf} based on the results reported in their paper, since there is no detailed experiment description and the models/codes are not available. As shown in \cref{HandNeRF_compare}, even if our model is trained from a single image, our model can capture the realistic texture of both hands comparably to HandNeRF. Note that HandNeRF is trained over hundreds of images from 10 views. Furthermore, BiTT is able to render in different illuminations as shown in \cref{Teaser} and \cref{MoreResult}.

\begin{table}[!ht]
\centering
\caption{Effects of coarse stage estimation, use of symmetric texture information (Sym. Tex.), and albedo consistency loss. Results are the mean value evaluated on novel poses and viewpoints. }
\label{ablation_table}
\begin{adjustbox}{scale=0.8}
\begin{tabular}{@{}ccc|lll@{}}
\toprule
Coarse Stage  & Sym. Tex. & $\mathcal{L}_{alb}$ & LPIPS↓ & PSNR↑ & SSIM↑ \\ \midrule
 \checkmark         &      &      &       0.1329&      25.36&      0.8940\\
 \checkmark         &   \checkmark   &      &  0.1230&      26.21&      0.9163\\
 \checkmark         &   \checkmark   &   \checkmark   &   \textbf{0.1176}    &      \textbf{27.16}&      \textbf{0.9199}\\ \bottomrule
\end{tabular}
\end{adjustbox}
\end{table}

\subsection{Ablation Study}
\paragraph{Symmetric information and Albedo Consistency Loss.} 
We perform an ablation study of the use of symmetric information in reconstructing hand texture. We replace the bi-directional connection with a uni-directional connection in BTR and $\mathcal{L}_{sym}$ is omitted. As shown in \cref{ablation_table}, not using the symmetric information significantly degrades the performance of reconstructing invisible side appearances.

We also perform an ablation study for the albedo consistency loss. In \cref{ablation_table}, the albedo consistency loss improves the overall performance of the model by more precisely estimating the albedo image.

\section{Conclusions}
In this work, we presented a novel two-hand texture reconstruction method from a single image called BiTT. First, the bi-directional texture reconstructor is proposed to create the full texture of both hands interactively. Second, we introduce a way to use the texture parametric model for recovering texture. The experimental results demonstrate that our method outperforms existing methods both qualitatively and quantitatively. We believe that our work can present a realistic experience to users by accurately representing their personalized hands in AR/VR applications.

\paragraph{Limitations and future work.}
One limitation is that although BiTT is robust to the geometric misalignments through the texture parametric model, there still exist instances where seriously misaligned meshes cause a significant level of noise. Future work could involve the learning method that can refine 3D meshes through detailed texture reconstruction. Another future work can be extending our system to detect and incorporate tattoos or accessories on the invisible side of the hands through generative networks\cite{LDM}, enhancing realism.

\paragraph{Acknowledgements.}
This work was in part supported by NST grant (CRC 21011, MSIT), KOCCA grant (R2022020028, MCST), IITP grant (RS-2023-00228996, MSIT). 

{
    \small
    \bibliographystyle{ieeenat_fullname}
    \bibliography{main}
}

\clearpage
\setcounter{page}{1}
\setcounter{section}{0}

\maketitlesupplementary

\renewcommand{\thesection}{\Alph{section}}

\section{Datasets}
\label{sec:datasets}

This section describes details of the datasets we used in the experiments. InterHand2.6M \cite{sup_InterHand2.6M} was mainly used and RGB2Hands \cite{sup_RGB2Hands} used as an additional dataset. For all datasets, the images are resized to 256 $\times$ 256 pixels.

\paragraph{InterHand2.6M \cite{sup_InterHand2.6M}.} 
InterHand2.6M is constructed by capturing sequential frames from multi-view videos of interacting two hands. It has 80 cameras for capturing subjects, which are 19 males and 7 females, in total 26 unique subjects. We selected camera viewpoints to ensure a diverse range of perspectives for the experiment. We used all 26 identities and randomly selected the poses along 17 ranges of motion (ROM) to demonstrate the fidelity of our model in various environments. 

\paragraph{RGB2Hands \cite{sup_RGB2Hands}.} 
We show the results of BiTT using the RGB2Hands dataset. Along with 4 different identities in the RGB2Hands dataset, we select a random image for training and select random other 40 images of different poses of the same identity. Note that RGB2Hands has low-resolution images and both hands generally show the same side of the hand (\cref{table3} in the main paper), thus it is less suited to demonstrate our method using the texture symmetry. The performance gain obtained is relatively less significant compared to that of InterHand2.6M. 

\section{Baselines}

\paragraph{S2Hand \cite{sup_S2Hand}.} We expanded the functionality of S2Hand, originally designed for single hand reconstruction, to accommodate both hands. We first increase the feature dimensions of the encoder to double times and use two separate texture regression layers for each hand. Informing symmetric information of both hand, the same mean texture color for each hand are initialized. We used ground truth hand mesh and pose for a fair comparison.

\paragraph{HTML \cite{sup_HTML}.} We utilize the principal components of the shadow-free version of the left and right hand. We used an HTML network in the coarse stage to estimate each hand texture vector. The texture vector is then multiplied by the corresponding principal components and generates a full hand UV map. For training the HTML network, we use the pixel-wise L1 reconstruction loss.

\paragraph{HARP \cite{sup_HARP}.}
HARP optimizes hand texture with visible pixel values from a monocular video. A monocular video including 4 frames of a hand was used to optimize HARP. InterHand2.6M \cite{sup_InterHand2.6M} has more than 300 points of the light source, showing a low rate of shadows. Thus, we remove the shadow rendering part in HARP and optimize the pixel values from the input sequence frames.

\begin{figure}[t]
  \centering
  \includegraphics[width=7.8cm]{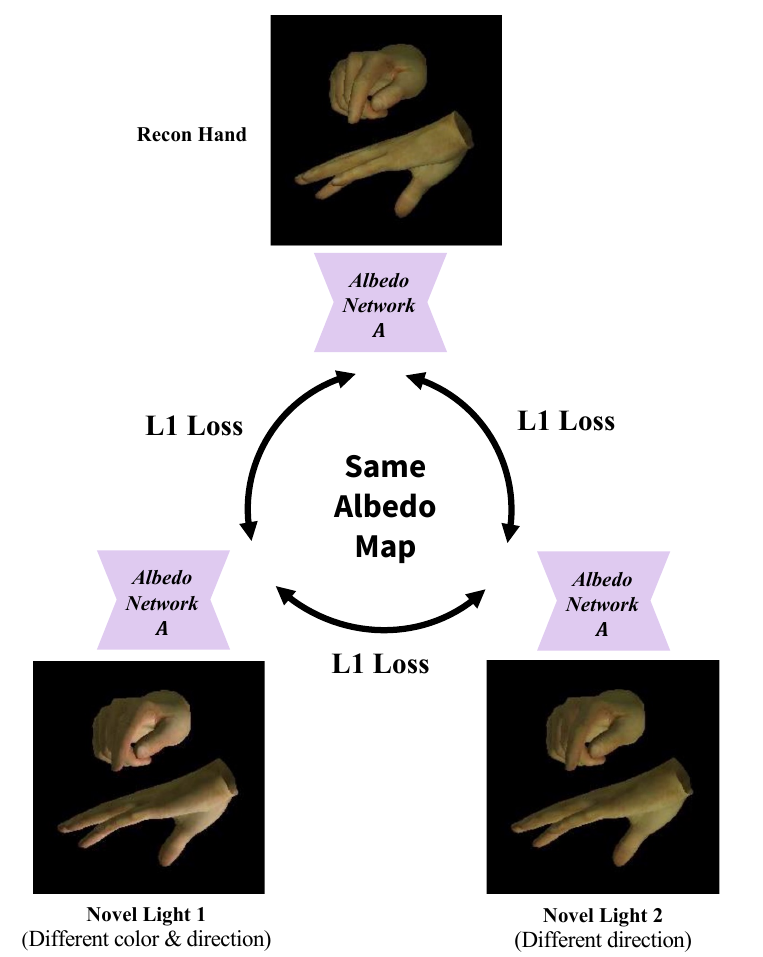}
  \caption{This figure shows the illustration of the albedo consisitency loss. }
  \label{AlbCons}
\end{figure}

\section{Implementation Details}
In this section, we describe a detailed implementation architecture. We used PyTorch for implementation. Our work is based on HTML \cite{sup_HTML} UV texture map template, where the UV map has the size of 1024 $\times$ 1024.

\begin{figure*}[!ht]
  \centering
  \includegraphics[width=17.3cm]{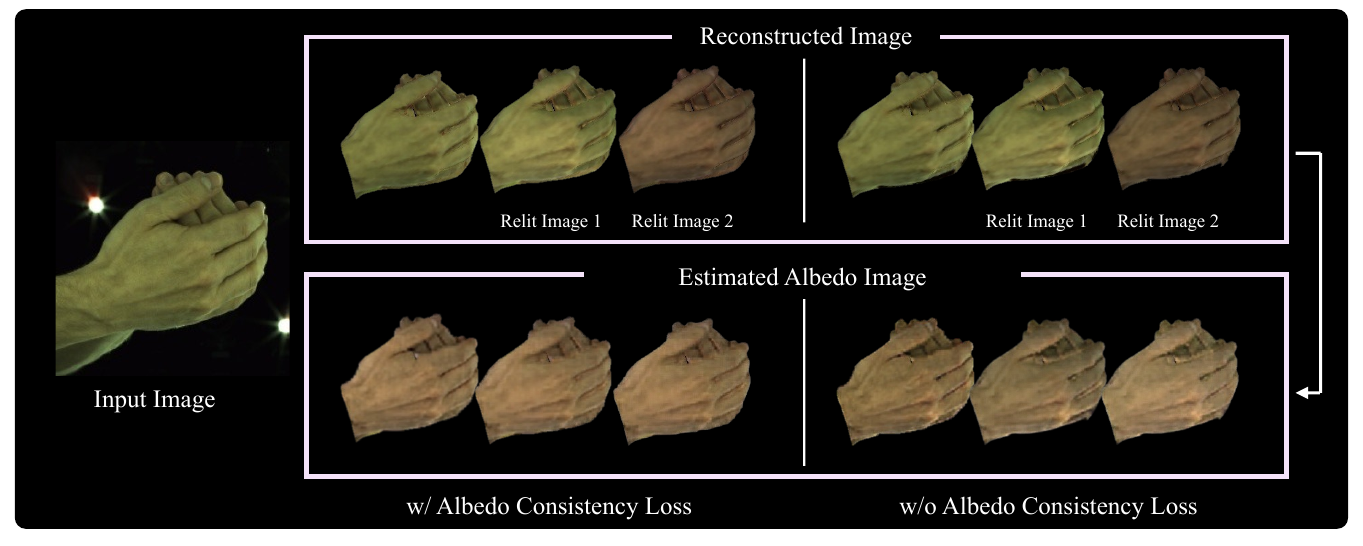}
  \caption{This figure shows the effectiveness of albedo consistency loss. In the process of reconstructing a hand from a given image and subsequently rendering it under two distinct lighting conditions, we generate three reconstructed and relit hand images. With three reconstructed/relightened hands, we estimate the albedo image for each. From the depicted images, we can find that utilizing albedo consistency loss can steadily estimate the albedo image, even when subjected to varying lighting conditions.}
  \label{AlbConsImg}
\end{figure*}

\paragraph{Albedo Network.} The albedo network directly follows the U-Net structure \cite{unet} where the input and output are an image. We use the LeakyReLU \cite{LeakyReLU} activation function in the encoding layer and the ReLU activation function in the decoding layers. In the albedo network, the encoding layer begins with input channels, which are progressively increased to 64, 128, 256, and 512 channels. After the encoding layers, the decoding layers reduce the feature dimension to 1024, 512, 256, and 128 channels, respectively. The reduction is achieved by concatenating the encoded features to the decoded features, while doubling the channel dimension. We upsample the feature by a scale factor of 2 with the nearest mode in each decoding layer. 

\paragraph{Light Network.} Light network uses an encoder network that takes an input image. The output of the light network which is denoted as $L$, is a 12-dimensional vector. This vector consists of different components: $L_{color}$, $L_{diff}$, $L_{spec}$, and $L_{direction}$, each of which is a 3-dimensional vector. The values for $L_{color}$, $L_{diff}$, and $L_{spec}$ are adjusted to between 0.2 and 1. We used ReLU for the activation function.

The light network $L$, starting from the input image, increases the feature dimension to 32, 64, 128, 256. Tanh is applied to the final activation function.

\paragraph{HTML Network.} HTML \cite{sup_HTML} network also uses an encoder network similar to the light network. The HTML network is used at the coarse stage to estimate HTML vectors for both hands. The HTML network output is a vector of $H(I) \in \mathcal{R}^{202}$, where $I$ is an input image and $H$ is an HTML network. Among the 202-dimensions, the first 101 features represent the left hand HTML vector $h_l$, while the last 101 features represent the right hand HTML vector $h_r$. Each vector is then multiplied to the corresponding principal components to generate a full UV map. Similar to the light network, we use the ReLU activation function for each encoding layer.

From the input image, the HTML network $H$ encodes the feature increasing the dimesion to 64, 128, 256, 512 and 202. Tanh is applied to the final activation function.

\paragraph{Bi-directional Texture Reconstructor (BTR).}
Bi-directional Texture reconstructor (BTR) is based on the ResNet \cite{resnet} structure. In each layer, it has 2 resblocks.  After the 2 sequence res blocks, we downsample the feature size into half. After each resblock, the feature dimension is increased to 16, 64, 128, and 256. The encoding network for each hand shares the same parameter weights.

The decoding part consists of a bidirectional decoding block. We have a total 3 decoding blocks with input feature dimensions of 48, 192, 384, and 768 ( 3 $\times$ (encoding channels)). We finally use the activation function sigmoid to ensure texture pixel colors between 0 and 1, preventing odd colors when rendered to 2D images.

\section{Albedo Consistency Loss}

The albedo consistency loss aims to ensure the consistency of estimated albedo maps across images rendered under different lighting conditions. \cref{AlbCons} shows an illustration of the albedo consistency loss concept. In this work, we have reconstructed the hand image and two other images that have been relit. Albedo network $A$ estimates the albedo map of each image. In our experiment, we constructed two different novel light conditions. Novel light 1 is characterized by a light source positioned at the bottom and has half the brightness of white light. On the other hand, novel light 2 differs from novel light 1 in terms of color, which is the same as the reconstructed light color. By applying the albedo consistency loss, we ensure that the albedo maps obtained from these three different light conditions remain consistent with each other. This helps to maintain the accurate representation of the object's reflectance properties regardless of the lighting variations.
 
\cref{AlbConsImg} illustrates the efficacy of the albedo consistency loss. It is shown that the albedo consistency loss allows a more precise and consistent estimation of the albedo image compared to not using it.

\begin{figure}[!ht]
  \centering
  \includegraphics[width=8.3cm]{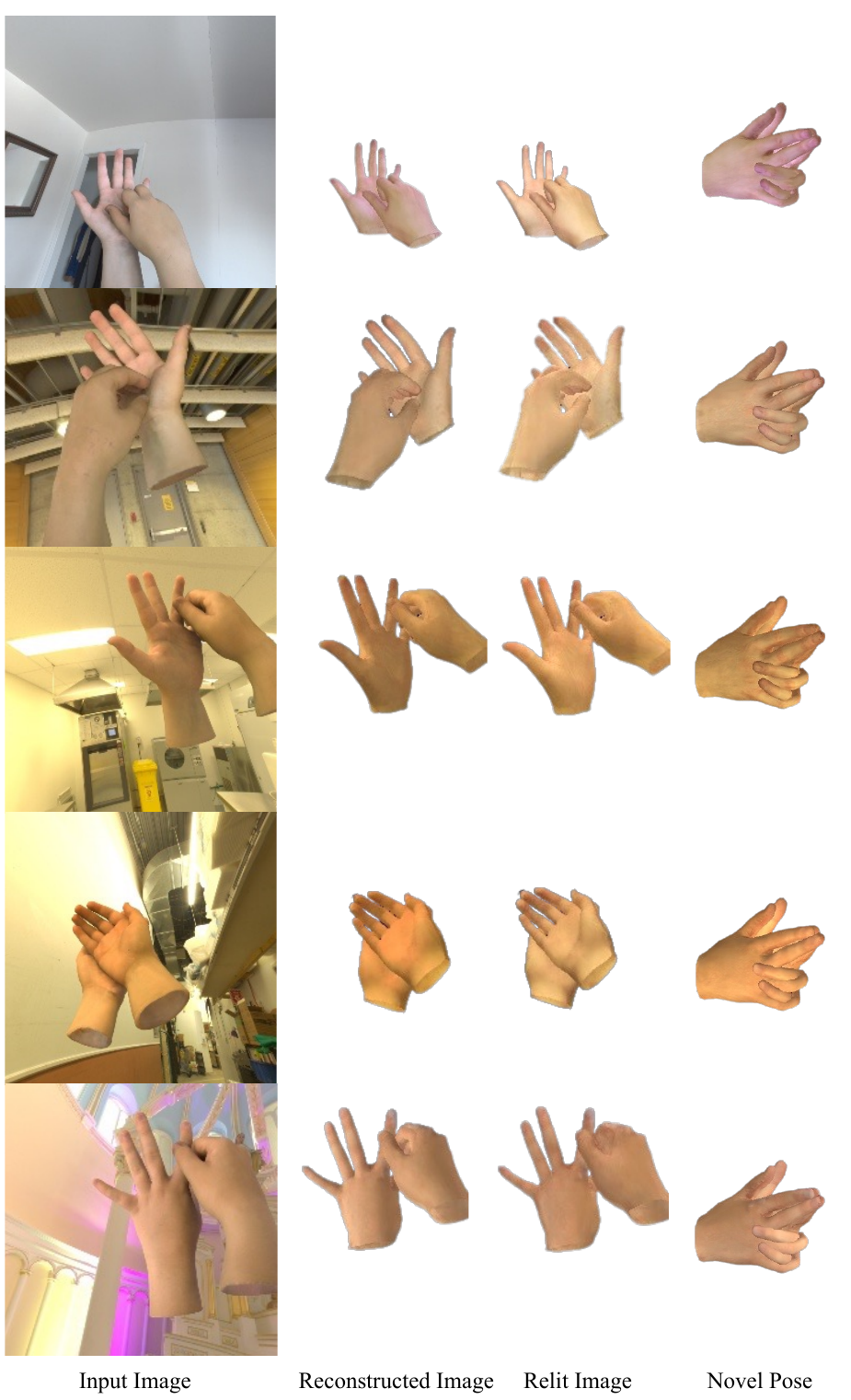}
  \caption{Qualitative results of BiTT on Re:InterHand \cite{Re-InterHand} dataset. We present a relightened image, and a novel pose of hands in the same lighting environment of input image. The identical novel pose images effectively show the differences of estimated lighting condition.}
  \label{suppresults1}
\end{figure}

\section{Subjective Tests}
We conducted a subjective test involving 27 users to answer 72 questions, earning a total of 1,944 responses. Each question presented a randomly generated image of each method with its corresponding GT image to evaluate the similarity and texture realism. Responses were collected in a 5-point discrete scale, ranging from "bad" (1) to "excellent" (5). 

\begin{figure}[h]
  \centering
  \includegraphics[width=8.3cm]{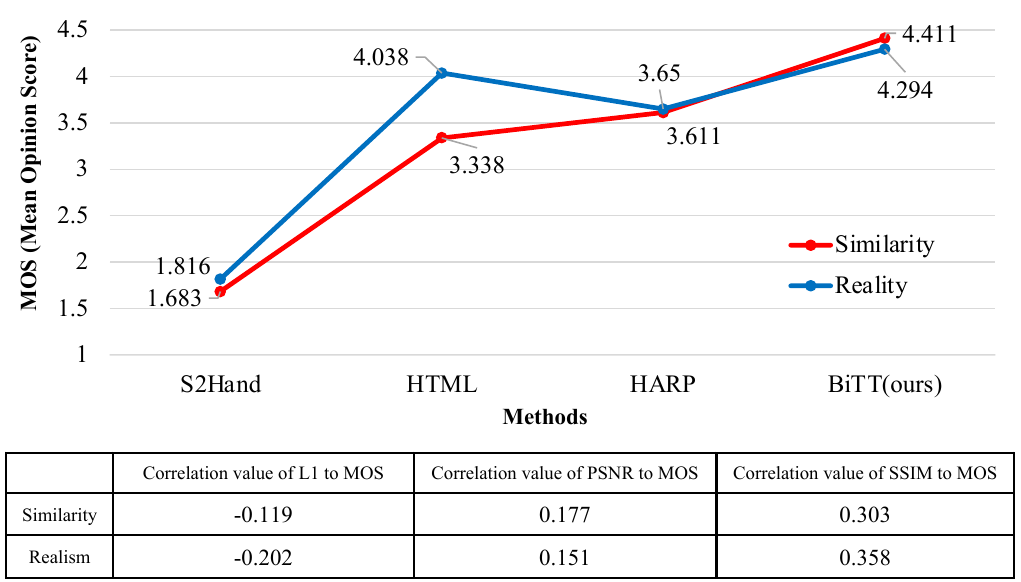}
  \caption{The subjective test results. }
  \label{SubTest}
\end{figure}

Our method achieved the best scores compared to baselines. S2Hand \cite{sup_S2Hand} showed lower scores due to blurred texture as a limitation of vertex rendering. HTML \cite{sup_HTML} showed strength in realism, while showing weakness in similarity. HARP \cite{sup_HARP} concluded with scores ranging between "fair" (3) and "good" (4), showing difficulty in the few shot learning. In \cref{SubTest}, we report subjective results with Pearson correlation value of the L1, PSNR, and SSIM metrics. 
\\[2pt]

\section{More Qualitative Results}

\begin{figure*}[!ht]
  \centering
  \includegraphics[width=16.7cm]{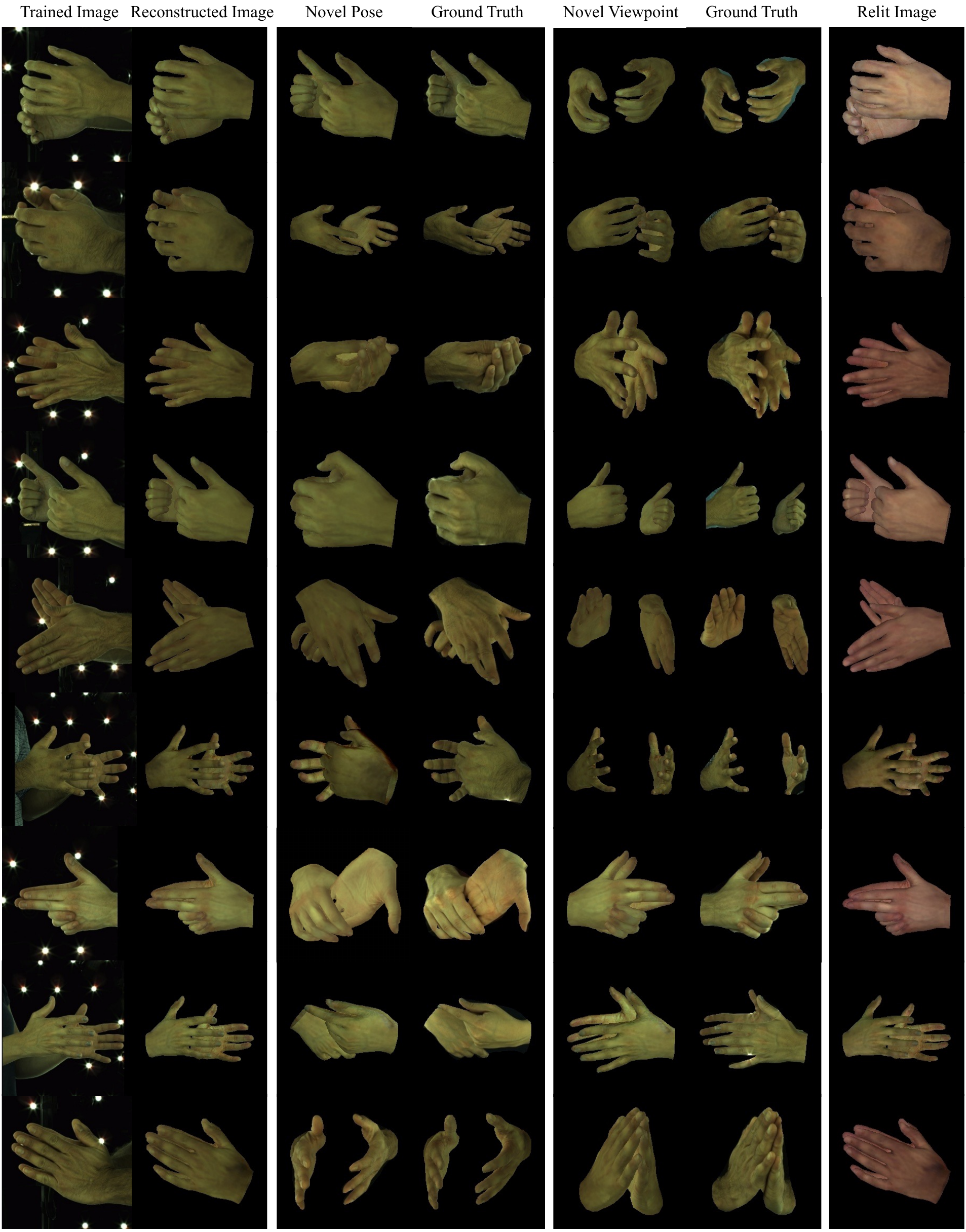}
  \caption{More qualitative results of reconstructed hands rendered on the novel pose, novel viewpoint, and relightened image in the dataset InterHand2.6M \cite{sup_InterHand2.6M}.}
  \label{suppresults2}
\end{figure*}

In this section, we present additional qualitative results of our reconstructed images of two hands, in \cref{suppresults2}. Even if the ground truth mesh does not perfectly align with the input hand image, our proposed method BiTT can present realistic hand textures. As a hand texture parametric model (HTML) \cite{sup_HTML} is robust to the noise and lacks its ability to represent background color, BiTT rendered image does not include background pixel colors on the hand texture due to minor geometric misalignments. Each column, from left to right, represents the input image, reconstructed image, novel pose, novel viewpoint with ground truth images, and relightened hands.

\paragraph{Results on Re:InterHand \cite{Re-InterHand} Dataset.}
Re:InterHand \cite{Re-InterHand} used the RelightableHands \cite{RelightableHands} method and generated a large dataset of two interacting hands relightened in several different environments.
RelightableHands presents the neural rendering approach to create relightable hand avatars. This process requires a specialized capturing studio having numerous cameras and light sources. Note our method only requires a single image input, end users in AR/VR systems can easily generate their personalized two-hand avatar realistically. 

As Re:InterHand has been passed through the learning process, the generated images exhibit less clear details of hands, such as wrinkles, hairs, and veins, compared to InterHand2.6M \cite{sup_InterHand2.6M} where the images are taken directly from cameras. We present several qualitative results of BiTT in the Re:InterHand dataset at \cref{suppresults1}. The figure demonstrates that BiTT accurately reconstructs a relightable two-hand avatar from an input image, even in diverse settings of environments.


\begin{figure}[!ht]
  \centering
  \includegraphics[width=8.3cm]{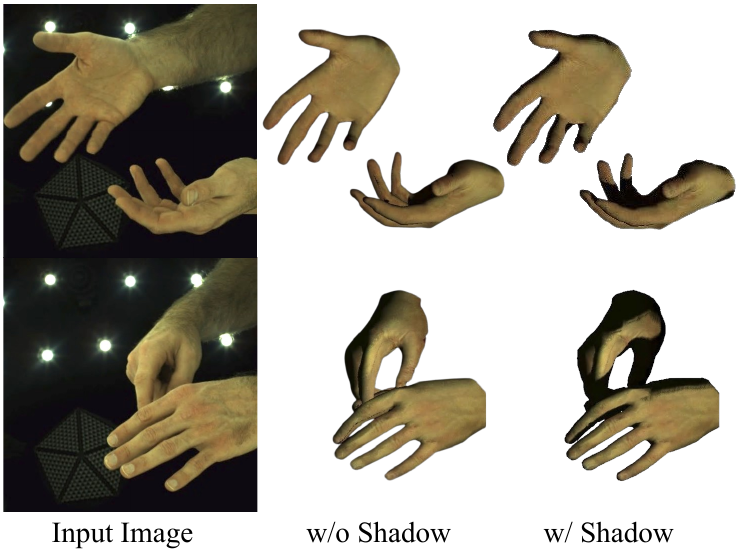}
  \caption{Visualization of the results with shadow rendering applied in BiTT method.}
  \label{shadow_applied}
\end{figure}
\vspace{-0.3cm}
\paragraph{Applying Shadow Construction.}
Our method relies on mesh-based rendering, which makes it easily compatible with traditional concepts in computer graphics. We have applied differentiable self-shadow rendering directly from the methodology presented in \cite{sup_HARP}. Given that our method involves two hands, it faces significant challenges in occlusion from each hand, along with self-occlusion toward the light source. Despite these complexities, it effectively captures the shadow appearance occured by self-occlusion and interhand-occlusion. The results are illustrated in \cref{shadow_applied}. Since InterHand2.6M \cite{sup_InterHand2.6M} was captured in an environment with numerous light sources, the application of shadow rendering tends to deviate from the input image. Nevertheless, in a scene illuminated by a single point light, it can enhance the realism of the rendered hand.

{
\small
~
}

\end{document}